# A Multimodal Visual Encoding Model Aided by Introducing Verbal Semantic Information

**Ma Shuxiao; Wang Linyuan; Yan Bin ***

(Henan Key Laboratory of Imaging and Intelligent Processing, PLA Strategic Support Force Information Engineering University, Zhengzhou 450001, China;)

mashuxiao23@163.com (S.M.); wanglinyuanwly@163.com (L.W.);
*Correspondence: ybspace@hotmail.com

**Abstract:** Biological research has revealed that the verbal semantic information in the brain cortex, as an additional source, participates in nonverbal semantic tasks, such as visual encoding. However, previous visual encoding models did not incorporate verbal semantic information, contradicting this biological finding. This paper proposes a multimodal visual information encoding network model based on stimulus images and associated textual information in response to this issue. Our visual information encoding network model takes stimulus images as input and leverages textual information generated by a text-image generation model as verbal semantic information. This approach injects new information into the visual encoding model. Subsequently, a Transformer network aligns image and text feature information, creating a multimodal feature space. A convolutional network then maps from this multimodal feature space to voxel space, constructing the multimodal visual information encoding network model. Experimental results demonstrate that the proposed multimodal visual information encoding network model outperforms previous models under the exact training cost. In voxel prediction of the left hemisphere of subject 1's brain, the performance improves by approximately 15.87%, while in the right hemisphere, the performance improves by about 4.6%. The multimodal visual encoding network model exhibits superior encoding performance. Additionally, ablation experiments indicate that our proposed model better simulates the brain's visual information processing.

## 1. Introduction

Vision, the primary source for humans to gather external information, is crucial to understanding how the human brain processes visual information. A critical approach to comprehending brain mechanisms is visual information encoding models, which simulate how the human visual cortex processes information to predict the response changes of different voxels under various external stimuli (Kay et al., 2008; Naselaris et al., 2011). Investigating neural encoding of visual information is essential in unraveling the brain's visual processing mechanisms and enhancing artificial visual models' perceptual and cognitive abilities.

In the philosophy of science, the collective human knowledge and perspectives on objects, attributes, and actions are called verbal information (Ivanova, n.d.). Through the visual system, humans can perceive various objects and actions in the natural world. Moreover, they can communicate and reason about these verbal categories through language and text. This suggests a connection between semantic information obtained through visual sensory input and language and text (Barsalou, 1999; Damasio, 1989; Ralph et al., 2017). Recent studies have revealed that verbal information is not solely localized to specific brain regions but distributed across the entire cortex (Anderson et al., 2016; Huth et al., 2016; Pereira et al., 2018; Xu et al., 2018). A graduate thesis from the Massachusetts Institute of Technology in 2022 (Ivanova, n.d.) explicitly mentioned that the brain's language regions are activated for nonverbal semantic tasks, such as image tasks. Researchers propose that in individuals with normal neural development, the brain may reencode the stimulus components of images into verbal forms as an additional source of task-relevant information (Connell & Lynott, 2013; Greene & Fei-Fei, 2014; Trueswell & Papafragou, 2010). As part of a control experiment, the researchers also conducted similar experiments with individuals with global aphasia. The results indicated that despite severe language impairments, these subjects could still participate in the experiment, albeit with lower efficiency than regular subjects. This indirectly confirms that the brain's cortical areas in regular subjects tend to transform image stimulus components into linguistic forms as an additional source of information (Ivanova, n.d.). Indeed, text-based computational models developed in recent years have successfully executed a wide range of " verbal " tasks, such as reasoning, paraphrasing, and question-answering tasks (Bao et al., 2022; Brown et al., 2020)。 In summary, text-based verbal information can provide a more comprehensive source of information for visual encoding within the brain's visual cortex, greatly enriching the "database" of visual information encoding models.

In the field of visual information encoding, whether it has the circular symmetric Gaussian difference-of-Gaussians (DoG) model improved upon by Zuiderbaan et al. in 2012(Zuiderbaan et al., 2012), the Bayesian population receptive field estimation model proposed by Zeidman et al. in 2018(Zeidman et al., 2018), or the "What" and "Where" models introduced by Wang et al. in 2021(Wang et al., 2021) based on hierarchical deep features of receptive fields, these models have undergone diverse improvements at various levels. However, the original input source for these models remains a single stimulus image.

In response to the current state of single-input visual information encoding models and inspired by the benefits of verbals in nonverbal tasks observed in modern biology, we propose a multi-modal visual information encoding model based on both stimulus image features and their related textual semantic features. Our model differs from traditional visual information encoding models in that while using the stimulus image features as the information source, we additionally introduce verbal information related to the stimulus image as linguistic semantic features. Subsequently, our model employs a Vision Transformer (Dosovitskiy et al., 2021) network to align image and text features, forming a multi-modal feature space. A CNN network then processes this space to predict the visual cortical voxel space, completing the mapping from stimulus image to visual cortical voxel.

Our work presents three key innovations:

1. The multi-modal visual information encoding model, for the first time, introduces textual semantic features as an additional information source to the encoding model instead of relying solely on stimulus images. This design closely resembles the processing pattern of visual information in the brain's visual cortical regions.

2. We utilize the Vision Transformer model to process multi-modal feature information, achieving alignment between features from different modalities.

3. Through cross-validation, we demonstrate significant performance improvements in our model compared to previous encoding models.

## 2. Methods

### 2.1 Model Overview

From the architectural perspective, the Multi-Modal Visual Information Encoding Model is an end-to-end framework. Its input consists of stimulus images and their corresponding verbal, while the output is the predicted voxel values generated by the encoding model. The multi-modal model comprises three components:

1. Graphical and Textual Feature Extraction Component: This component involves extracting graphical and textual features.

2. Multi-Modal Information Interaction Component: This part involves the interaction between graphical and textual features. It is a crucial step for integrating information from both modalities.

3. Feature-Voxel Mapping Component: This component is common in traditional encoding tasks and maps features to voxel responses.

Following a single-stream approach, the Multi-Modal Visual Information Encoding Model employs the Vision Transformer (ViT) as the backbone network. ViT-B/32 model is utilized, with a hidden size of 768, 12 layers, patch size 32, 3072 MLP size, and 12 self-attention heads.

Regarding textual features, diverse types of textual information are considered. Different pre-trained models, including BERT and GPT-2, are employed to process distinct textual features. These models are referred to as Textual Feature Extraction Models.

The information processing flow of the multi-modal model is as follows: First of all, The input stimulus image $I_{img} \in \mathbb{R}^{C \times H \times W}$ is sliced and flattened into a vector $v_{img} \in \mathbb{R}^{N \times (P^2 \cdot C)}$, where $N$ is $HW / P^2$. Followed by linear projection $V \in \mathbb{R}^{(P^2 \cdot C) \times H}$ and position embedding $V^{pos} \in \mathbb{R}^{(N+1) \times H}$, $v_{img}$ is embedded into $\overline{v}_{img} \in \mathbb{R}^{N \times H}$. For textual information, due to varying lengths of descriptive text between different images, we adopt a fixed-length approach to constrain the length of textual information. We use an empirical value of L=256 for the text length. The input text $t_{text} \in \mathbb{R}^{L \times |V|}$ is embedded to $\overline{t}_{text} \in \mathbb{R}^{L \times H}$ with a word embedding matrix $T \in \mathbb{R}^{|V| \times H}$ and a position embedding matrix $T^{pos} \in \mathbb{R}^{(L+1) \times H}$。 The text ($\overline{v}_{img} \in \mathbb{R}^{N \times H}$) and image ($\overline{t}_{text} \in \mathbb{R}^{L \times H}$) embeddings are summed with their corresponding modal-type embedding vectors $t^{type}, v^{type} \in \mathbb{R}^{H}$, then are concatenated into a combined sequence $z^0$. The contextualized vector $z$ is iteratively updated through D-depth transformer layers up until the final contextualized sequence $z^D$. $p$ is a pooled representation of the whole multimodal input, and is obtained by applying linear projection $W_{pool} \in \mathbb{R}^{H \times H}$ and hyperbolic tangent upon the first index of sequence $z^D$. Subsequently, we pass the vector $z^D$ through a CNN convolutional network for feature reduction. Then, we use FC layers to complete the mapping from multi-modal feature information to voxel responses. This process generates the model's predicted voxel values. The overall architecture diagram of the multi-modal visual information encoding model is illustrated below:

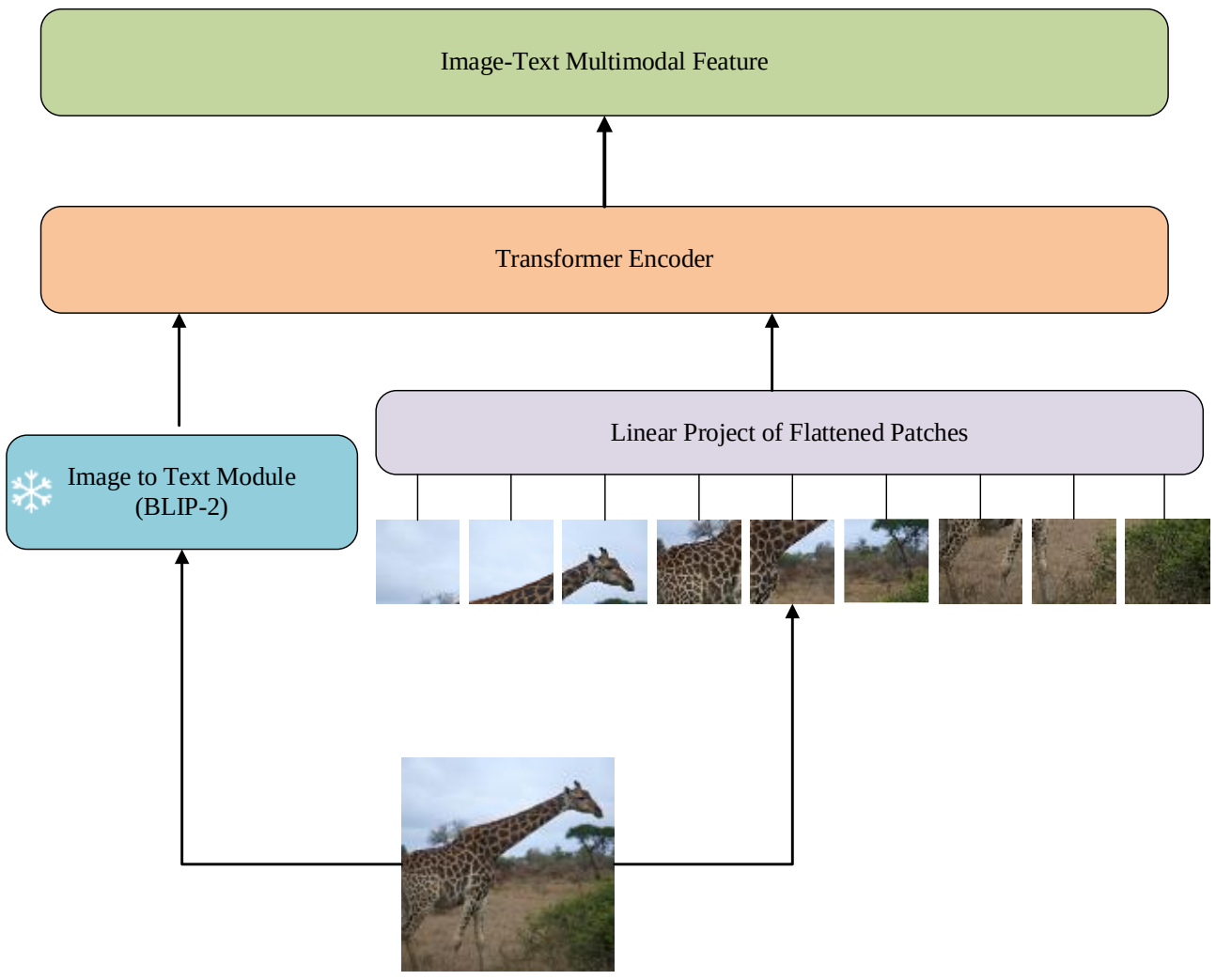

A

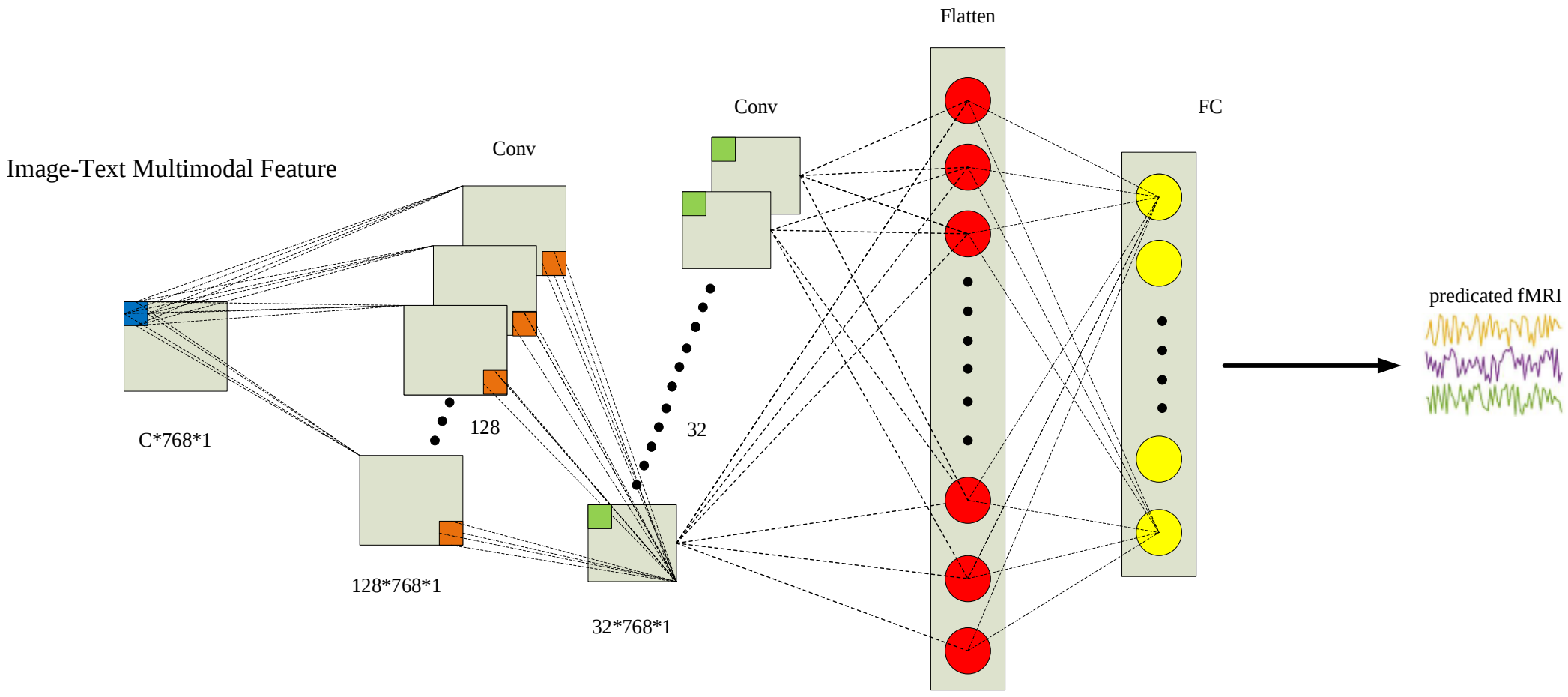

B

**Figure 1.** The overall architecture diagram of the multi-modal visual information encoding model. Figure 1A is the graphical and verbal feature extraction component; Figure 1B is the feature-voxel mapping component.

## 2.2 Train Objectives

Researchers commonly employ the Pearson correlation coefficient as an evaluation metric for the visual encoding model. Therefore, to closely align with the final evaluation metric in this experiment, we also utilize the Pearson correlation coefficient as the loss function.

The formula for the Pearson correlation coefficient is as follows:

$$R_v = \mathrm{corr}(G_v, P_v) = \\ = \frac{\sum_t \left(G_{v,t} - \bar{G}_v\right)\left(P_{v,t} - \bar{P}_v\right)}{\sqrt{\sum_t \left(G_{v,t} - \bar{G}_v\right)^2 \sum_t \left(P_{v,t} - \bar{P}_v\right)^2}},$$

where $v$ is the index of vertices (over all subjects and hemispheres), $t$ is the index of the test stimuli images, $G$ and $P$ correspond to, respectively, the ground truth and predicted fMRI test data, $\bar{G}$ and $\bar{P}$ are the ground truth and predicted fMRI test data averaged across test stimuli images, $R$ is the Pearson correlation coefficient between $G$ and $P$.

## 3. Experiments

### 3.1 Dataset

This experiment is based on the largest and richest dataset of fMRI responses to natural scenes, the Natural Scenes Dataset (NSD). Please visit the NSD website for more details. Briefly, NSD provides data acquired from a 7-Tesla fMRI scanner over 30–40 sessions during which each subject viewed 9,000–10,000 color natural scenes (22,000–30,000 trials)(Allen et al., 2022). We analyzed data for four of the eight subjects who completed all imaging sessions (subj01, subj02). The images used in the NSD experiments were retrieved from MS COCO and cropped to 224*224. The training sets for these two subjects each contain 9841 images. The test sets also consist of 159 images, respectively.

The fMRI data is z-scored at each NSD scan session and averaged across image repeats, resulting in 2D arrays with the amount of images as rows and as columns a selection of the vertices that showed reliable responses to images during the NSD experiment. The left (LH) and right (RH) hemisphere files consist of, respectively, 19,004 and 20,544 vertices, with the exception of subjects 6 (18,978 LH and 20,220 RH vertices) and 8 (18,981 LH and 20,530 RH vertices) due to missing data.

Previous research has revealed that the visual cortex is divided into multiple distinct areas having different functional properties, referred to here as regions-of-interest (ROIs). Some of those are functionally defined, for example by their preferred response to a particular category such as faces or houses. Others are defined by anatomical criteria. Following is the list of ROIs (ROI class file names in parenthesis):

Early retinotopic visual regions (prf-visualrois): V1v, V1d, V2v, V2d, V3v, V3d, hV4.

Body-selective regions (floc-bodies): EBA, FBA-1, FBA-2, mTL-bodies.

Face-selective regions (floc-faces): OFA, FFA-1, FFA-2, mTL-faces, aTL-faces.

Place-selective regions (floc-places): OPA, PPA, RSC.

Word-selective regions (floc-words): OWFA, VWFA-1, VWFA-2, mfs-words, mTL-words.

Anatomical streams (streams): early, midventral, midlateral, midparietal, ventral, lateral, parietal.

## 3.2 Implementation Details

For all experiments, we use AdamW(Loshchilov & Hutter, 2019) optimizer with base learning rate of $10^{-4}$ and weight decay of $10^{-2}$. The learning rate is set to decay by 0.8 every 5 epochs to ensure the model learns more effectively. We resized the input images to 224x224 while maintaining the aspect ratio. For the ViT-B/32 model, this generates 49 image patches. To ensure accuracy in our experiments, we employed a 5-fold cross-validation approach throughout our study.

The stimulus images in the NSD dataset are, in fact, a subset of the Microsoft COCO dataset. Each image in the COCO dataset is accompanied by five text descriptions, which multiple annotators independently provide. Recognizing the diversity of these text descriptions, we employed the CLIP model in our experiments to select the best-matching textual information for each natural image from the five available descriptions. This selected text information was used as the relevant textual information for our stimulus images in this study.

We utilized pre-trained models from the Hugging Face repository to achieve better initialization of model parameters and reduce the training cost. Specifically, we employed the publicly available BERT model for our study.

## 3.3 Experimental Results

Due to the large size of the NSD dataset, to enhance experimental efficiency and ensure result accuracy, we randomly selected data from both hemispheres of subjects 1 and 2, resulting in a total of 4 experimental data sets. We performed 5-fold cross-validation to validate the experimental data.

Table 1 presents the predictions of the four datasets in various ROIs using our model. It is worth noting that there are significant variations in predicted values among different voxels within each ROI. To mitigate the impact of extreme values, in this experiment, we use the median value of all voxels within each ROI to showcase our model's voxel predictions.

**Table 1** Median voxel predictions across various ROIs for the four datasets are as follows:

| | Subject 1 | | Subject 2 | |
|---|---|---|---|---|
| | LH | RH | LH | RH |
| Early | 0.142042 | 0.239547 | 0.189492 | 0.289508 |
| Midventral | 0.192652 | 0.232347 | 0.265559 | 0.298260 |
| Midlateral | 0.202285 | 0.256350 | 0.257363 | 0.303731 |
| Midparietal | 0.296695 | 0.245398 | 0.178395 | 0.280645 |
| Ventral | 0.304879 | 0.233720 | 0.416443 | 0.288365 |
| Lateral | 0.418475 | 0.261894 | 0.373886 | 0.325633 |
| Parietal | 0.328863 | 0.262444 | 0.268750 | 0.309305 |
| All vertices | 0.262446 | 0.246072 | 0.290660 | 0.299426 |

Figures 2A, 2B, 2C, and 2D depict the results of the four datasets using both the traditional encoding model (with only stimulus images as input, indicated by blue markers) and our multimodal model (indicated by red markers).

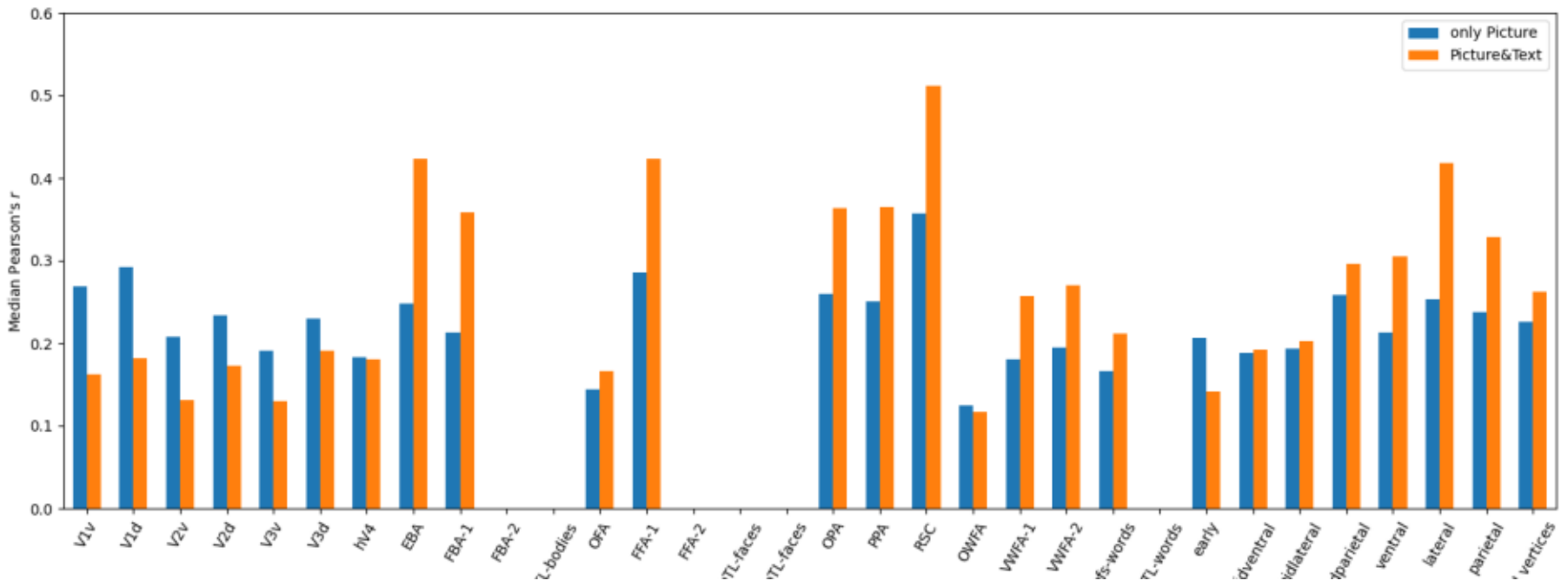

A

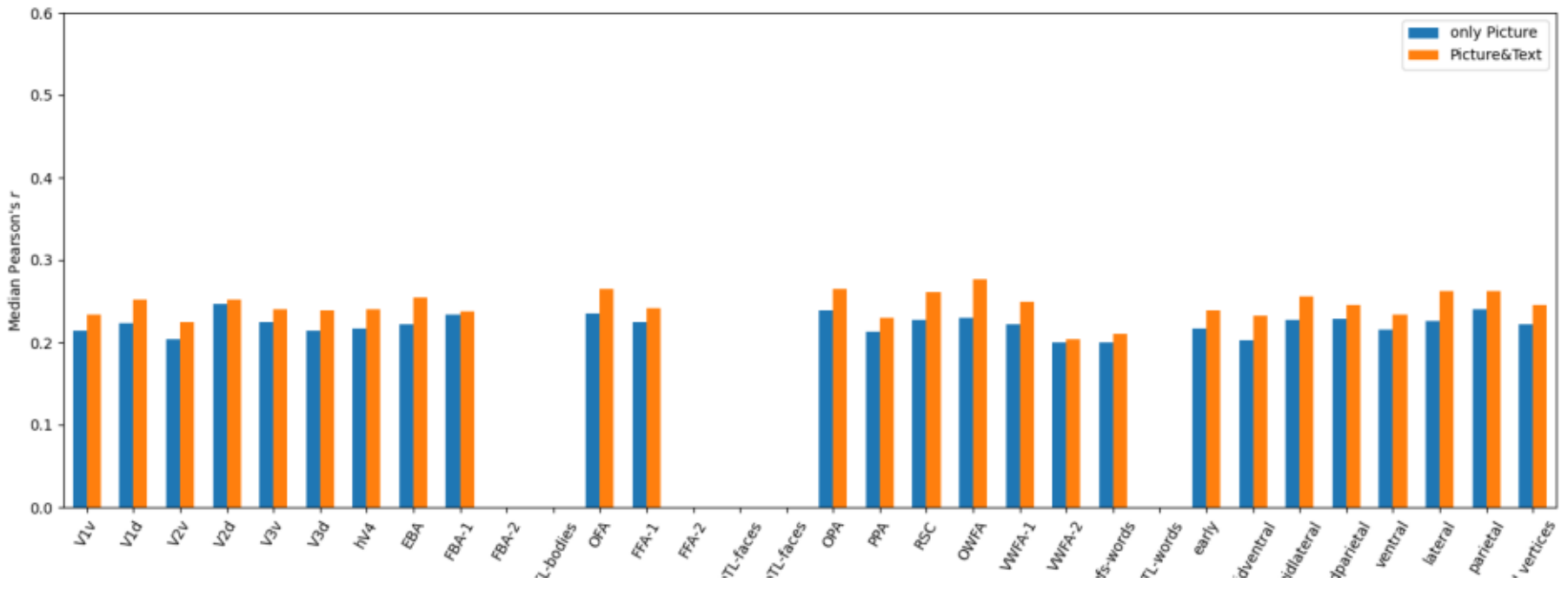

B

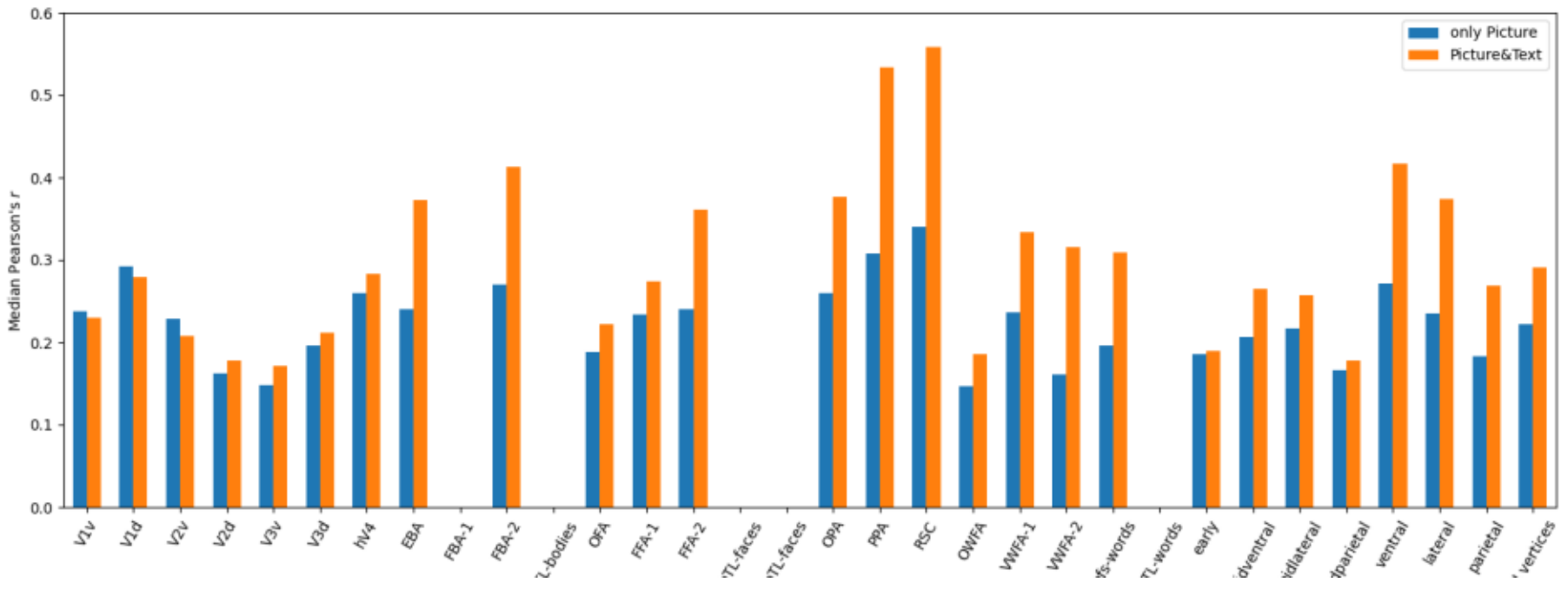

C

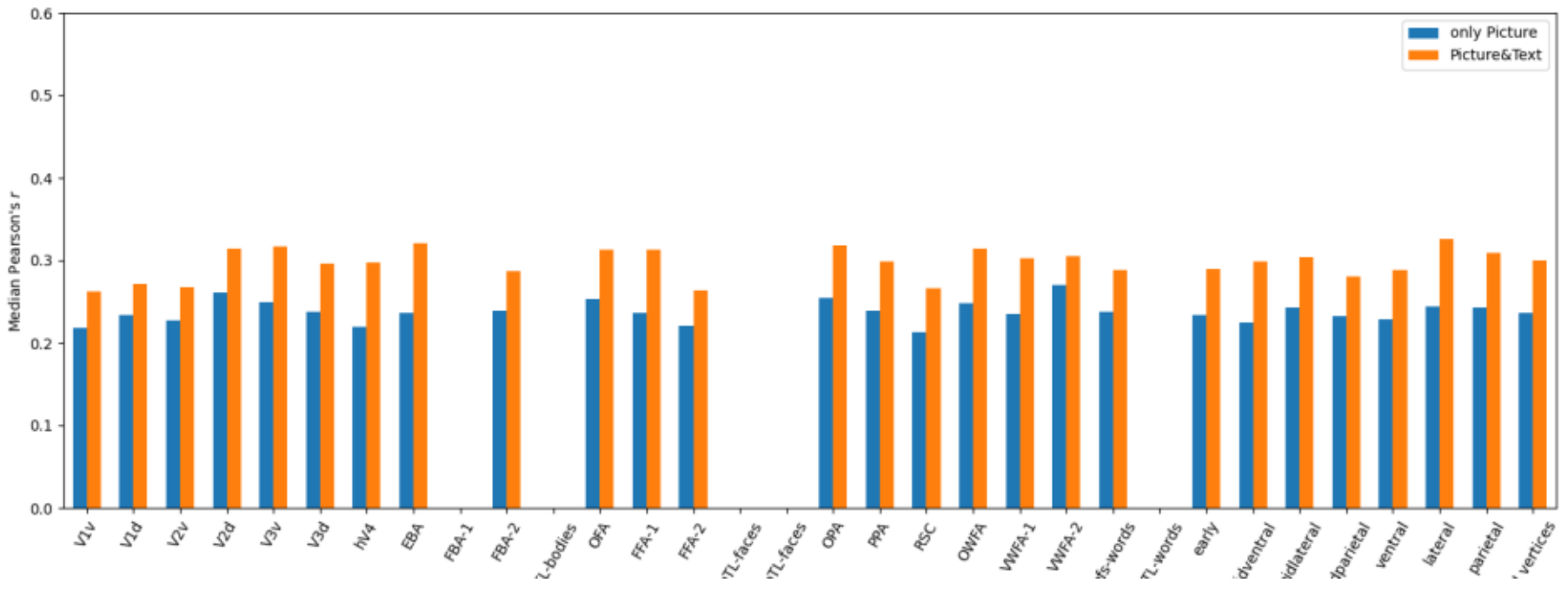

D

**Figure 2.** The comparison of encoding performance between our model and the traditional model. Figures 2A and 2B respectively illustrate the comparison of encoding performance between our model and the traditional model in the left and right hemispheres of subject 1; Figures 2C and 2D respectively illustrate the comparison of encoding performance between our model and the traditional model in the left and right hemispheres of subject 2.

Comparing Figure 2A and Figure 2C, we observe that in both the left hemisphere data of Subject 1 and Subject 2, our model demonstrates improved encoding accuracy compared to the traditional encoding model that employs a single stimulus image as input source. Our model exhibits significantly better encoding accuracy starting from higher-level regions like hV4. For instance, in the PPA and RSC regions, the encoding accuracy is improved by 73.35% and 64.24%, respectively. However, in the lower-level visual areas V1-V3, the traditional encoding model has a slight advantage in Subject 1's left hemisphere, while this advantage is not prominent in Subject 2's left hemisphere.

Upon comparing Figure 2B and Figure 2D, it is evident that the performance of both models remains consistent in the right hemisphere data of both Subjects. Specifically, our model consistently outperforms the traditional model regarding encoding performance.

### 3.4 Ablation Study

Building upon the experiments in Section 3.3, we conducted additional ablation experiments to illustrate further the advantages of textual features in the multi-modal network model.

In Section 3.4, the textual information used was derived from the original COCO dataset's descriptions. However, human-generated descriptions can vary significantly due to individual interpretations of stimulus images. We employed a text-image generation model to mitigate this variability to generate textual information for the stimulus images. This text-image generation model is based on the pre-trained ViT-GPT2 model in the Hugging Face community. In this setup, ViT serves as the image decoder, and GPT2 is utilized for textual encoding. For a detailed description of the ViT-GPT2 model, please refer to this website: https://huggingface.co/nlpconnect/vit-gpt2-image-captioning.

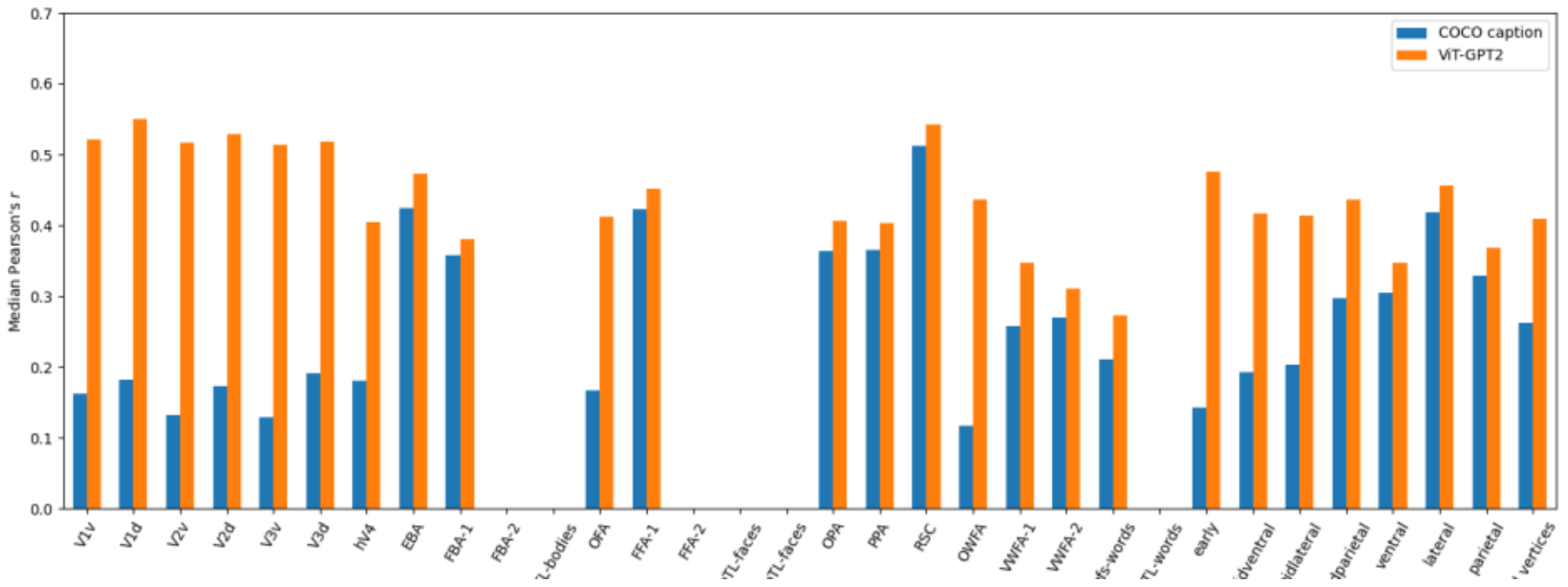

**A**

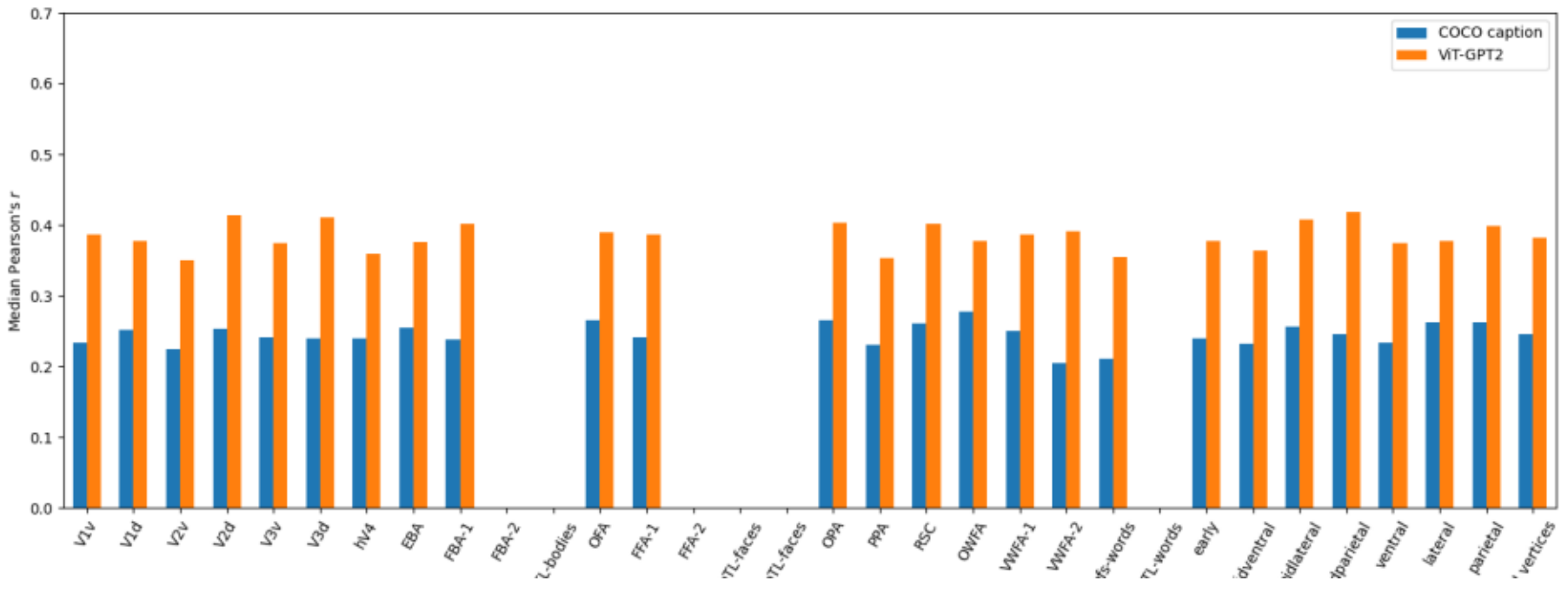

**B**

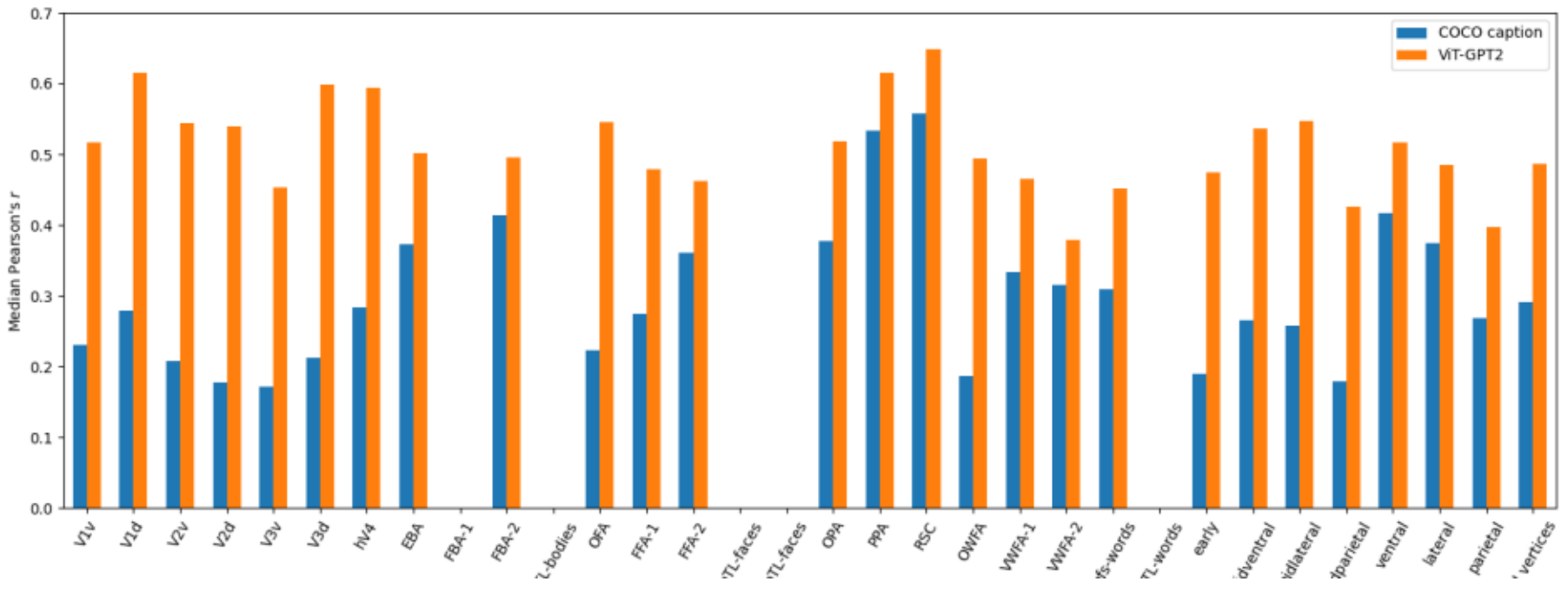

**C**

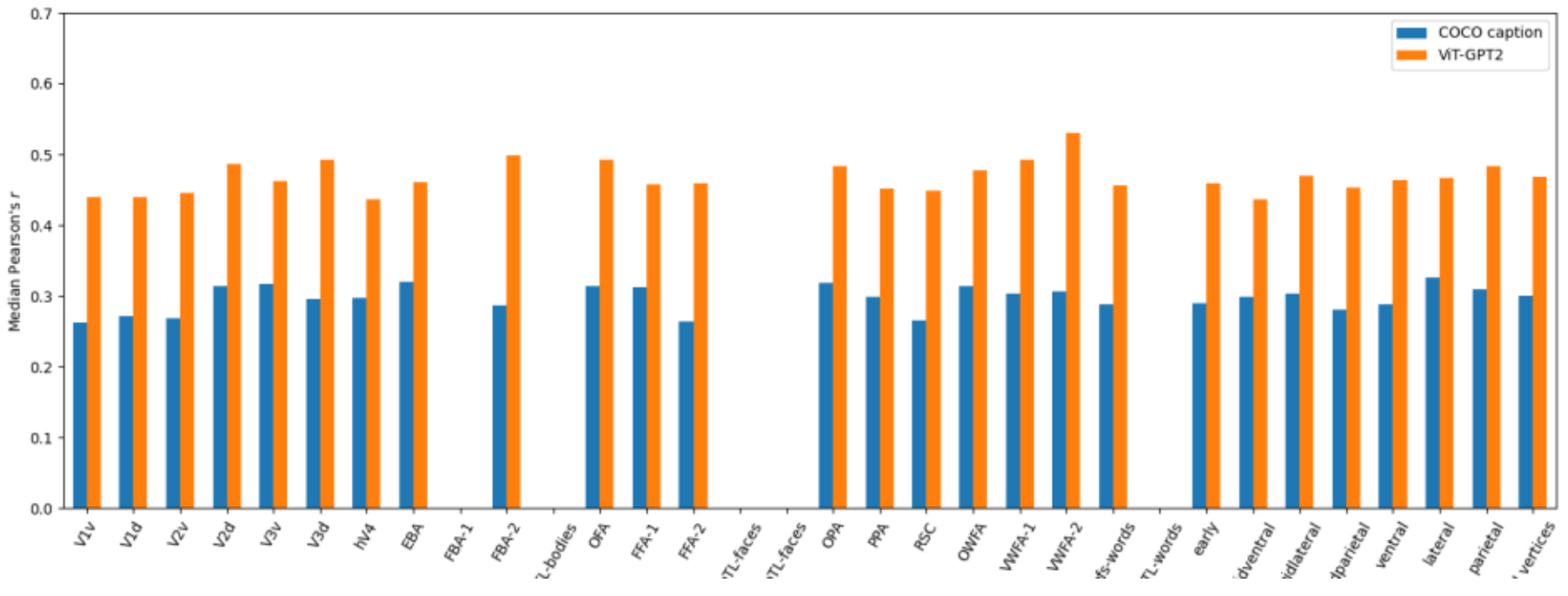

**D**

**Figure 3.** Comparison of performance between the text generated by the COCO dataset and the image-text generated model output in our model. Figures 3A and 3B respectively depict the performance comparison of the left and right hemispheres of subject 1; Figures 3C and 3D illustrate the performance comparison of the left and right hemispheres of subject 2.

From Figure 3A, 3B, 3C and 3D, it is evident that the four sets of test data show significant improvement in the text-image generation model. This observation suggests that the performance of our encoding model is reliant on the quality of the textual information.

**Table 2.** Comparison of Text Descriptions from Different Sources. This table describes the best-matched descriptions from the COCO dataset and the descriptions generated by the ViT-GPT2 model for a randomly selected set of stimulus images from the training dataset.

| | COCO | ViT-GPT2 |
|---|---|---|
| 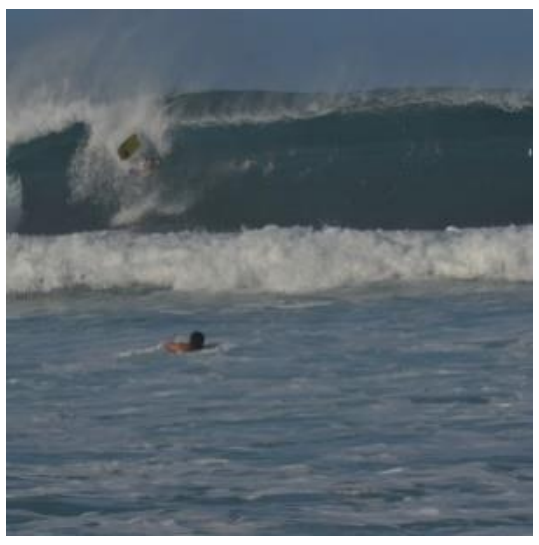 | A person is swimming in the water near a wave. | a person riding a surfboard on top of a wave. |
| 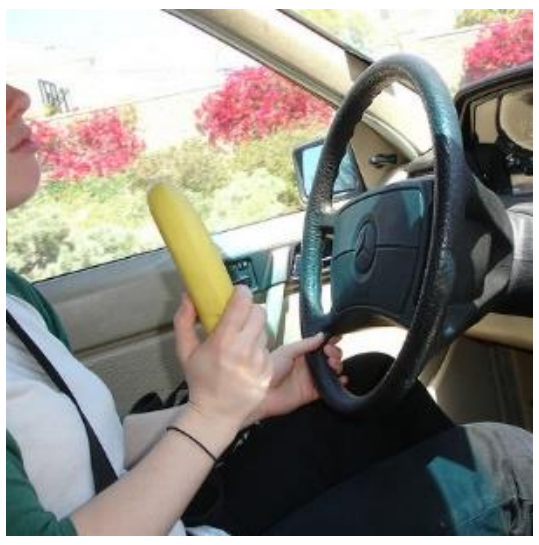 | A driver holds a banana with greater care than the steering wheel of the car she is operating. | a woman holding a banana in her hand. |
| 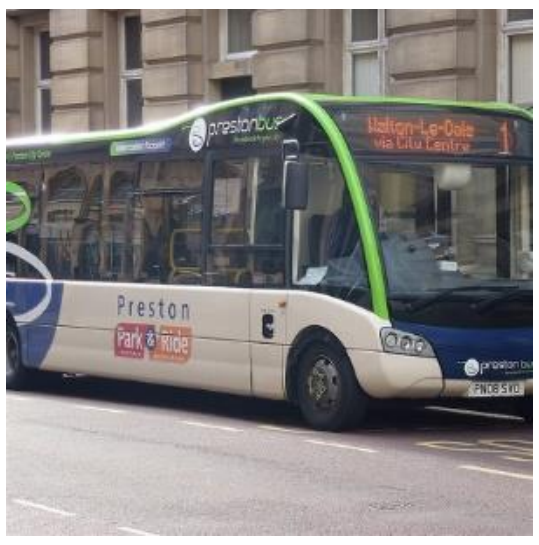 | A park and ride bus is on the street. | a bus is parked on the side of the road. |
| 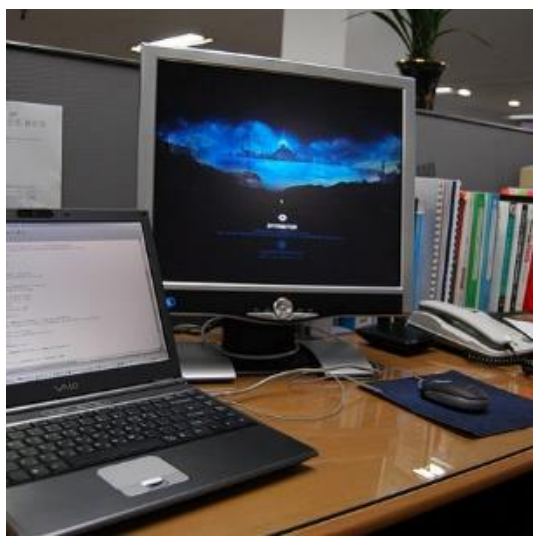 | A shiny looking computer desk with a laptop and monitor. | a desk with a laptop and a monitor. |

In Table 2, we randomly selected 4 images from the training dataset. We displayed the best-matched descriptions from COCO's description using the CLIP model and the descriptions generated using our ViT-GPT2 model. The descriptions generated by ViT-GPT2 are more specific and provide more straightforward descriptions of the main objects in the images.

As mentioned earlier, text information is considered an additional source that participates in non-verbal semantic tasks in the visual cortex. Similarly, for our visual encoding task, is text information also an "additional" source of information? To address this question, we retrained the encoding model using only stimulus images and extended the training iterations appropriately.

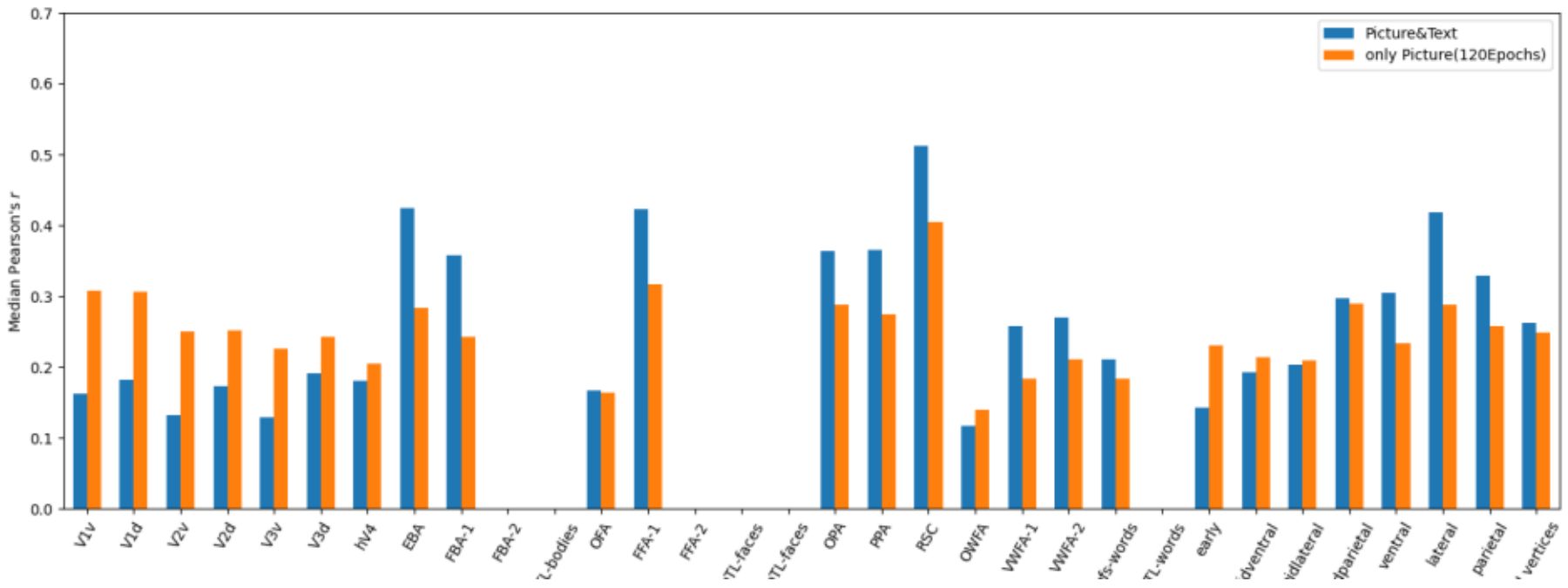

Figure 4A

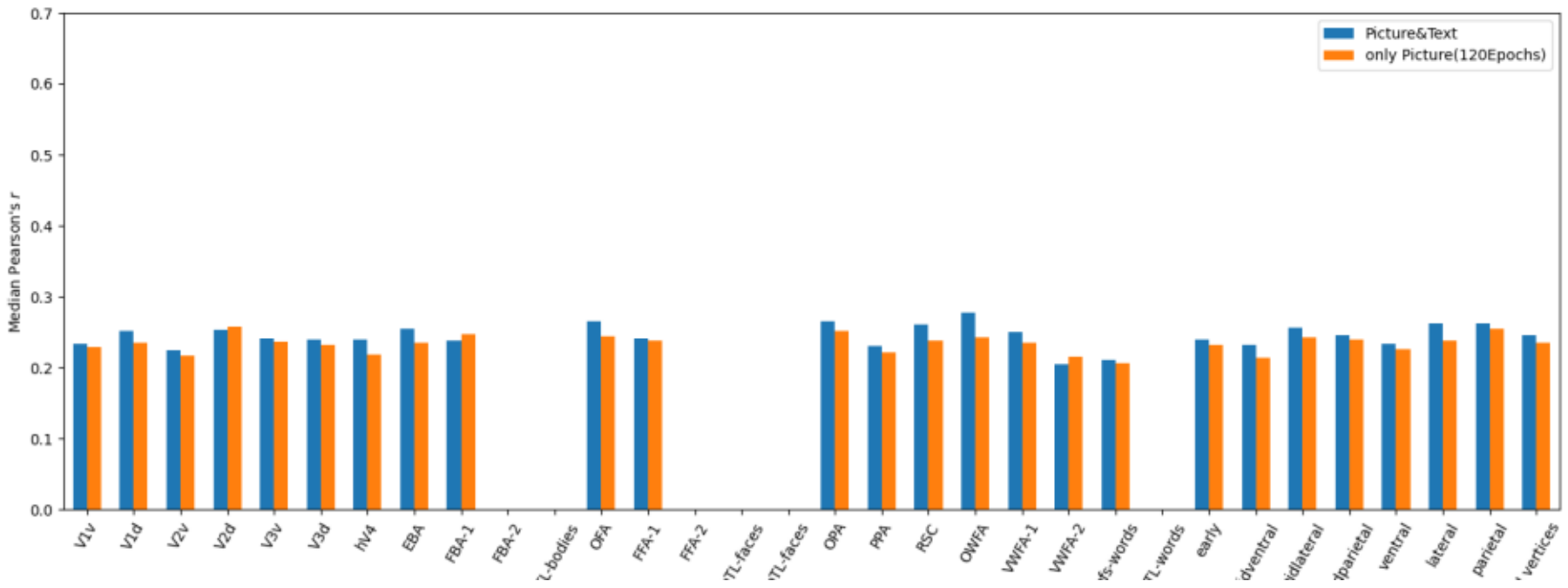

Figure 4B

In this ablation experiment, we focused on using only stimulus images as input information sources. We adjusted the number of epochs from 30 to 120 while keeping other experimental parameters and model hyperparameters unchanged. In Figure 4, we observe the comparison between data from the left and right brain of Subject 1. In Figure 4A, the performance of the 120 Epochs experiment is dominant in the early stages of the visual coding, up to the hV4 region, and shows a reversed trend beyond the hV4 region. The performance tends to align in the "All Vertices" category. This trend is even more pronounced in the right brain data of Subject 1, where the performance of 120 Epochs and the multi-modal model becomes consistent.

Through this ablation experiment, we confirm that textual features as non-linguistic semantic features in visual coding models are indeed extra information from linguistic semantic tasks. Suppose we do not introduce this kind of additional information in the laboratory, although we can achieve a certain level of coding effect by extending the training time. In that case, it increases the time and training cost. By incorporating textual information as an additional information source, we can significantly reduce training costs and time, enabling coding models with equivalent parameter settings to achieve better performance at lower training costs.

## 4. Conclusion and Future Work

In this paper, inspired by fundamental biological research highlighting the involvement of text semantic information as an additional source in visual information processing, we introduce a multi-modal visual information encoding network model based on both image feature information and verbal semantic information. This model thoroughly considers the role of textual information in the brain's visual processing. We align verbal feature information with image data through Transformer networks, integrating text semantic information as an extra information source into the visual encoding model. This approach allows the model to closely mimic the brain's visual information processing pattern.

Experimental results demonstrate that our proposed multi-modal network model outperforms traditional single-modal models in terms of performance, even at the same training cost. Crucially, our model presents a novel encoding paradigm for the future of visual coding. By introducing aligned textual feature information, we enhance the prominence of text semantic information in the visual encoding model. This encourages the model to achieve better encoding performance with lower training costs. This contribution opens up a new avenue for the field of visual coding, suggesting that incorporating aligned textual feature information could greatly improve encoding performance while minimizing training expenses. This could enable a broader and more comprehensive understanding of this field.